\documentclass{article}

\usepackage{microtype}
\usepackage{adjustbox}
\usepackage{graphicx}
\usepackage{subcaption}
\usepackage{booktabs} 

\usepackage{url}

\usepackage{breakurl}
\usepackage[breaklinks]{hyperref}

\usepackage{siunitx}
     
\usepackage{amsmath}

\usepackage[accepted]{icml2020}

\icmltitlerunning{How to Train Your Neural ODE: the World of Jacobian and
Kinetic Regularization}

\usepackage{mathtools}

\usepackage{amssymb} 
\usepackage{amsthm}
\usepackage{url}

\usepackage{siunitx}
\usepackage{graphicx}
\usepackage{float}
\usepackage{multirow}
\graphicspath{{./figs/}}

\newtheorem{theorem}{Theorem}[section]

\theoremstyle{definition}

\theoremstyle{remark}
\newtheorem{remark}[theorem]{Remark}

\newcommand{\norm}[1]{\Vert#1\Vert}

\newcommand{\abs}[1]{\vert#1\vert}

\renewcommand{\div}[1]{\operatorname{div} \left( #1 \right)}
\DeclareMathOperator{\grad}{\nabla}

\DeclareMathOperator{\KL}{{KL}}
\DeclareMathOperator{\tr}{{tr}}

\DeclareMathOperator*{\E}{\mathbb{E}}
\newcommand{\bq}{\begin{equation}}
\newcommand{\eq}{\end{equation}}

\newcommand{\mb}[1]{\mathbf #1}
\newcommand{\R}{\mathbb{R}}
\newcommand{\Rd}{\R^d}

\newcommand{\trp}{\mathsf{T}}

\begin{document}

\twocolumn[
\icmltitle{How to Train Your Neural ODE: the World of Jacobian and
Kinetic Regularization}



\icmlsetsymbol{equal}{*}

\begin{icmlauthorlist}
\icmlauthor{Chris Finlay}{mc}
\icmlauthor{J\"orn-Henrik Jacobsen}{vec}
\icmlauthor{Levon Nurbekyan}{la}
\icmlauthor{Adam M Oberman}{mc}
\end{icmlauthorlist}

\icmlaffiliation{mc}{Department of Mathematics \& Statistics, McGill University,
Montr\'eal, Qu\'ebec, Canada}
\icmlaffiliation{vec}{Vector Institute, University of Toronto, Toronto, Ontario, Canada}
\icmlaffiliation{la}{Department of Mathematics, UCLA, California, USA}

\icmlcorrespondingauthor{Chris Finlay}{christopher.finlay@mcgill.ca}

\icmlkeywords{Machine Learning, ICML}

\vskip 0.3in
]



\printAffiliationsAndNotice{}  

\begin{abstract}
Training neural ODEs on large datasets has not been tractable due to the
necessity of allowing the adaptive numerical ODE solver to refine its step size
to very small values. In practice this leads to dynamics equivalent to many
hundreds or even thousands of layers. In this paper, we overcome this apparent
difficulty by introducing a theoretically-grounded combination of both optimal
transport and stability regularizations which encourage neural ODEs to prefer
simpler dynamics out of all the dynamics that solve a problem well. Simpler
dynamics lead to faster convergence and to fewer discretizations of the solver,
considerably decreasing wall-clock time without loss in performance. Our
approach allows us to train neural ODE-based generative models to the same
performance as the unregularized dynamics, with significant reductions in
training time. This brings neural ODEs closer to practical relevance in
large-scale applications.
\end{abstract}

\section{Introduction}
\begin{figure}[ht]
  \vskip -0.1in
    \centering
    \begin{subfigure}[t]{.45\columnwidth}
      \includegraphics[height=3in]{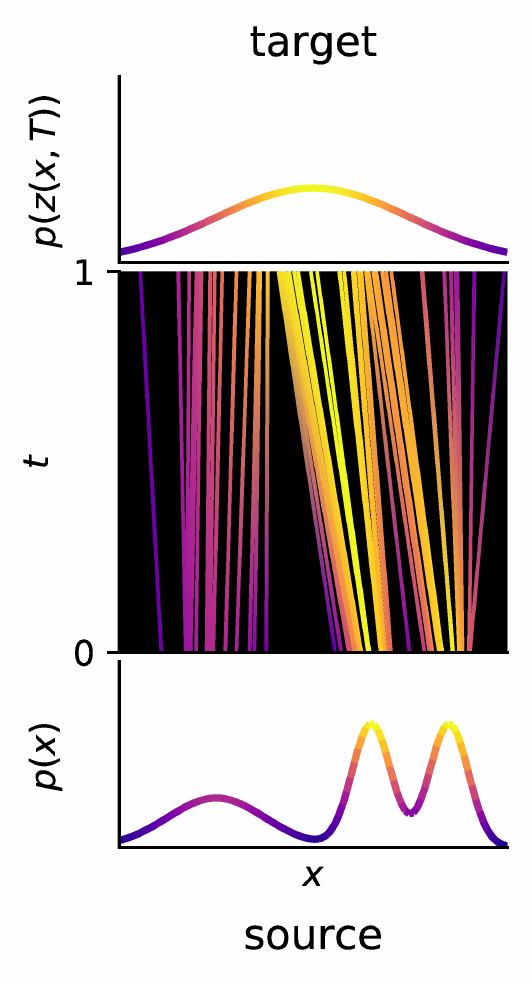}
  \caption{Optimal transport map}\label{fig:ot}
  \end{subfigure}
  \hspace{1em}
  \begin{subfigure}[t]{.45\columnwidth}
    \includegraphics[height=3in]{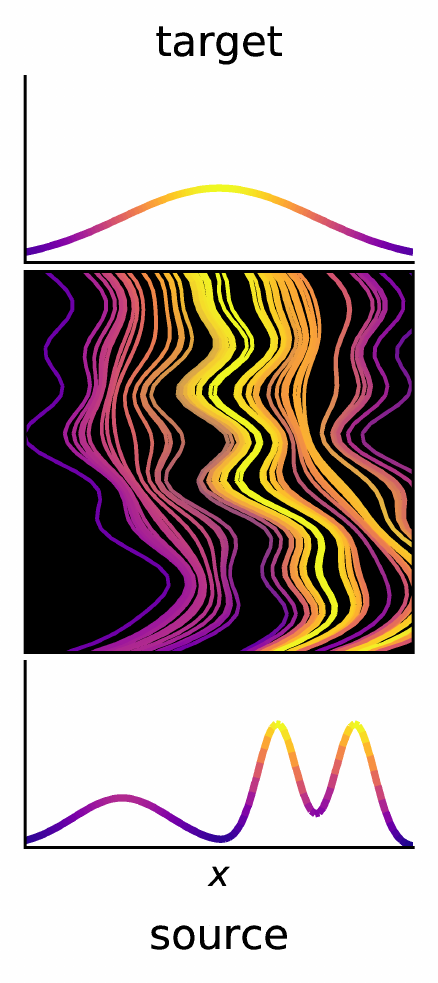}
  \caption{generic flow}\label{fig:squiggles}
  \end{subfigure}
  \caption{Optimal transport map and a generic normalizing flow.}\label{fig:1}
\vskip -0.1in
\end{figure}
Recent research has bridged dynamical systems,
a workhorse of
mathematical modeling, with neural networks, the defacto function approximator
for high dimensional data. The great promise of this pairing is that the vast
mathematical machinery stemming from dynamical systems can be leveraged for
modelling high dimensional problems in a dimension-independent fashion.

Connections between neural networks and ordinary differential equations (ODEs) were almost
immediately noted after residual networks  \cite{resnet} were first proposed. Indeed, it was observed that there is a striking
similarity between ResNets and the numerical solution of ordinary differential
equations
\citep{e_proposal_2017,haber2017stable,ruthotto2018deep,chen2018neural,behrmann2018invertible}.
In these works, deep networks are interepreted as discretizations of an
underlying dynamical system, where time indexes the ``depth" of the network and the
parameters of the discretized dynamics are learned. An alternate viewpoint was
taken by neural ODEs \citep{chen2018neural}, where the dynamics of the neural
network are approximated by an adaptive ODE solver on the fly. This latter
approach is quite compelling as it does not require specifying the number of layers of the network
beforehand. Furthermore, it allows the
learning of homeomorphisms without any structural constraints on the function computed
by the residual block. 

Neural ODEs have shown great promise in the physical sciences 
\citep{kohler2019equivariant}, in modeling irregular time series
\citep{rubanova2019latent}, mean field games \cite{ruthotto19}, continuous-time
modeling \cite{YildizHL19,kanaasimple}, and for generative modeling through normalizing flows with free-form Jacobians \citep{grathwohl_ffjord}.
Recent work has even adapted neural ODEs to the stochastic setting
\cite{li2020sde}.
Despite these successes, some hurdles still remain. 
In particular, although neural ODEs are memory efficient, they can take a
prohibitively long time to train, which is arguably one of the main stumbling
blocks towards their widespread adoption.

In this work we reduce the 
training time of neural ODEs by regularizing the learned dynamics, complementing
other recent approaches to this end such as augmented neural ODEs \cite{augmentednodes}.
Without further constraints on their dynamics, high dimensional neural ODEs may
learn dynamics which minimize an objective function, but which generate
irregular solution trajectories. See for example Figure \ref{fig:squiggles}, where an
unregularized flow exhibits undesirable properties due to unnecessarily fluctuating dynamics.
As a solution, we propose two theoretically motivated regularization terms arising from an optimal transport viewpoint of the learned map, which
encourage well-behaved dynamics (see \ref{fig:ot} left). We empirically demonstrate that proper regularization leads to significant speed-up in training time without loss in performance, thus bringing neural ODEs closer to deployment
on large-scale datasets. Our methods are validated on the problem of generative
modelling and density estimation, as an example of where neural ODEs have shown impressive results, but could easily be applied elsewhere.

In summary, our proposed regularized neural ODE (RNODE) achieves the same performance as the baseline, while reducing the wall-clock training time by many hours or even days.

%

\section{Neural ODEs \& Continuous normalizing flows}
Neural ODEs simplify the design of deep neural networks by formulating the
forward pass of a deep network as the solution of a ordinary differential
equation.  Initial work along these lines was motivated by the similarity of the
evaluation of one layer of a ResNet and the Euler discretization of an ODE.
Suppose the block in the $t$-th layer of a ResNet is given by the function $\mb
f(\mb x, t; \theta)$, where $\theta$ are the block's parameters. Then the
evaluation of this layer of the ResNet is simply $\mb x^{t+1} = \mb x^t + \mb f(\mb x^t, t; \theta)$.
Now, instead consider the following ODE
\begin{equation}\label{eq:dynamics}\
  \begin{cases} \dot {\mb {z}} = \mb f(\mb z, t; \theta) \tag{ODE}\\
    \mb z(0) = \mb x
  \end{cases}
\end{equation}
The Euler discretization of this ODE with step-size $\tau$ is $\mb z^{t+1} = \mb
z^t + \tau \mb f(\mb z^t, t; \theta)$, which is nearly identical to the forward
evaluation of the ResNet's layer (setting step-size $\tau=1$ gives equality).
Armed with this insight, \citet{chen2018neural} suggested a method  for training
neural networks based on \eqref{eq:dynamics} which abstain
from \emph{a priori} fixing step-size. \citeauthor{chen2018neural}'s method is a continuous-time
generalization of residual networks, where the dynamics are generated by an
\emph{adaptive} ODE solver that chooses step-size on-the-fly.

Because of their adaptive nature, neural ODEs can be more flexible than ResNets in certain scenarios, such as when trading between model speed and accuracy. Moreover given a fixed
network depth, the memory footprint of neural ODEs is orders of magnitude
smaller than a standard ResNet during training. They therefore show great
potential on a host
of applications, including generative modeling and density estimation. 
An apparent drawback of neural ODEs is their long training time: although a
learned function $\mb f(\cdot\,; \theta)$ may generate a map that solves a problem
particularly well, the computational cost of numerically integrating
\eqref{eq:dynamics} may be so prohibitive that it is not tractable in practice. 
In this paper we demonstrate this need not be so: with proper regularization, it is possible to learn $\mb
f(\cdot\,; \theta)$ so that \eqref{eq:dynamics} is easily and quickly solved.
\begin{figure*}[ht]
  \vskip 0.2in
    \centering
    \begin{subfigure}[t]{.45\textwidth}
      \includegraphics[height=2.5in]{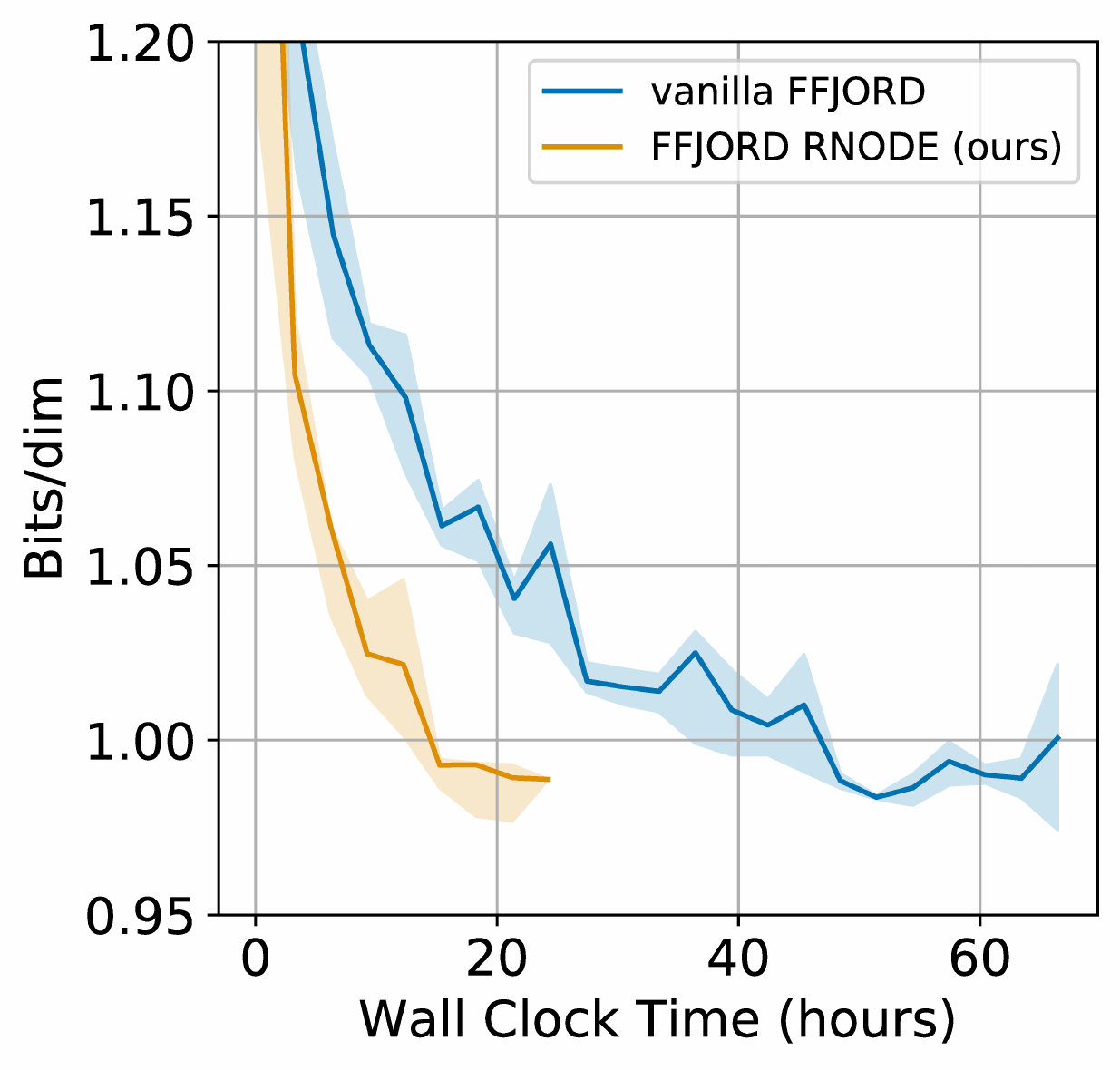}
    \caption{MNIST}\label{fig:bpd-mnist}
    \end{subfigure}
    \begin{subfigure}[t]{.45\textwidth}
      \includegraphics[height=2.5in]{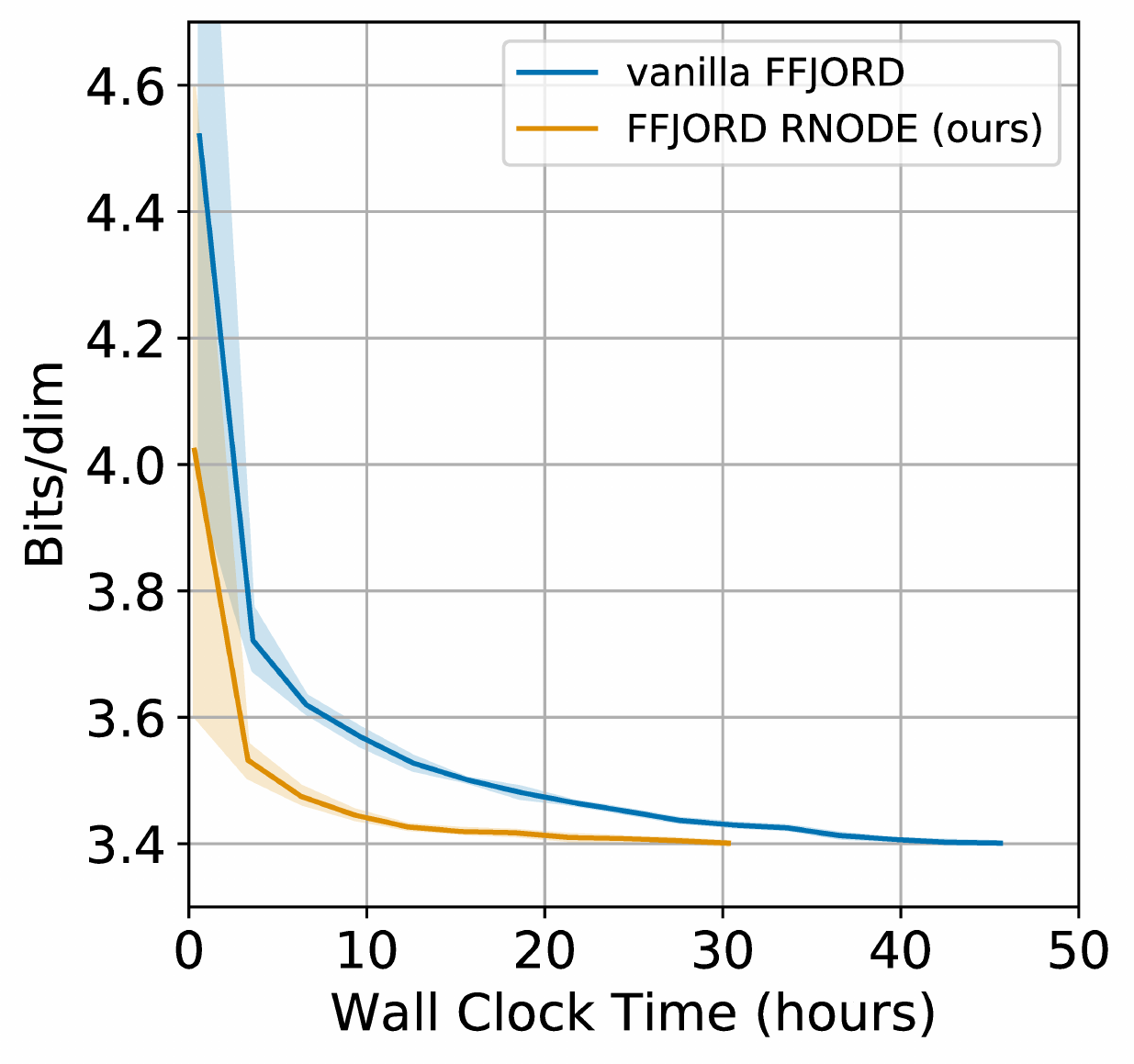}
    \caption{CIFAR10}\label{fig:bpd-cifar10}
  \end{subfigure}
  \caption{Log-likelihood (measured in bits/dim) on the validation set as a function of wall-clock time. Rolling
average of three hours, with 90\% confidence intervals.}\label{fig:bpd}
\end{figure*}

\subsection{FFJORD}
In density estimation and generative modeling, we wish to estimate an unknown data distribution $p(\mb x)$ from which
we have drawn $N$
samples. Maximum likelihood seeks to approximate $p(\mb x)$ with a parameterized
distribution $p_{\theta}(\mb x)$ by minimizing the Kullback-Leibler divergence between the two, or
equivalently minimizing
\begin{equation}\label{eq:opt1}
  J(p_{\theta}) =  -\frac{1}{N}\sum_{i=1}^N \log p_{\theta}(\mb x_i) 
\end{equation}
Continuous normalizing flows \citep{grathwohl_ffjord,chen2018neural} parameterize $p_{\theta}(\mb x)$ using a vector field
$\mb f: \Rd\times \R \mapsto \Rd$ as follows. 
Let $\mb z(\mb x, T)$ be the solution map given by running the dynamics
\eqref{eq:dynamics}
for fixed time $T$.
Suppose we are given a \emph{known} distribution $q$ at final time $T$,
such as the normal distribution. Change of variables tells us that the
distribution $p_{\theta}(\mb x)$ may be evaluated through
\begin{equation}\label{eq:change}
  \log p_{\theta}(\mb x) = \log q\left(\mb z(\mb x, T)\right) + \log \det \abs{\grad \mb z(\mb
  x,T)}
\end{equation}
Evaluating the log determinant of the Jacobian is difficult. 
\citet{grathwohl_ffjord} exploit the following identity 
from fluid mechanics \citep[p 114]{villani2003topics}
\begin{equation}
  \frac{\partial}{\partial t} \log \det \abs{\grad\mb z(\mb
x,t)} = \div{ \mb f} (\mb z(\mb x,t),t)) 
\end{equation}
where $\div{\cdot}$ is the divergence operator\footnote{In the normalizing flow literature divergence is typically written explicitly as the trace of the Jacobian, however we use $\div{\cdot}$ which is more common elsewhere.}, $\div{\mb f}(\mb x) = \sum_i
\partial_{x_i} f_i(\mb x)$.
By the fundamental theorem of calculus, we may then rewrite \eqref{eq:change} in integral form
\begin{equation}\label{eq:changeint}
  \log p_{\theta}(\mb x) = \log q\left(\mb z(\mb x, T)\right) + \int_0^T\div{ \mb f} (\mb
  z(\mb x,s),s) \, \mathrm d  s
\end{equation}
\begin{remark}[Divergence trace estimate]In \citep{grathwohl_ffjord}, the divergence
  is estimated using an unbiased Monte-Carlo trace
estimate \citep{hutchinson1990stochastic,avrom11}, 
\begin{equation}\label{eq:divtrest}
  \div{\mb f}(\mb x) = \E_{\mb \epsilon \sim \mathcal N(0,1)} \left[\epsilon^\trp \grad
  \mb f(\mb x) \mb \epsilon \right ]
\end{equation}
\end{remark}
By using the substitution \eqref{eq:changeint}, the task of maximizing log-likelihood shifts from
choosing $p_{\theta}$ to minimize \eqref{eq:opt1}, to learning the flow generated by
a vector field $\mb f$. This results in a normalizing flow with a free-form Jacobian
and
reversible dynamics, and was
named FFJORD by \citeauthor{grathwohl_ffjord}.

\subsection{The need for regularity}
The vector field learned through FFJORD that maximizes the
log-likelihood is not unique, and raises troubling problems related to the
regularity of the flow. For a simple example, refer to Figure \ref{fig:1}, where
we plot two normalizing flows, both mapping a toy one-dimensional distribution
to the unit Gaussian, and where both maximize the log-likelihood of exactly the
same sample of particles. Figure \ref{fig:ot} presents a ``regular'' flow, where
particles travel in straight lines that travel with constant speed. In contrast,
Figure \ref{fig:squiggles} shows a flow that still maximizes the log-likelihood,
but that has undesirable properties, such as rapidly varying local trajectories
and non-constant speed.

From this simple motivating example, the need for regularity of the vector field
is apparent. Without placing demands on the vector field $\mb f$, it is entirely
possible that the learned dynamics will be poorly conditioned. This is not just
a theoretical exercise: because the dynamics must be solved with a numerical
integrator, poorly conditioned dynamics will lead to difficulties during
numerical integration of \eqref{eq:dynamics}. Indeed, later we present results
demonstrating a clear correlation between the number of time steps an adaptive
solver takes to solve \eqref{eq:dynamics}, and the regularity of $\mb f$.

How can the regularity of the vector field be measured?
One motivating approach is to measure the force experienced by a particle $\mb z(t)$ under the
dynamics generated by the vector field $\mb f$, which is given by the total derivative
of $\mb f$ with respect to time
\begin{align}
  \frac{\mathrm d \mb f(\mb z, t)}{\mathrm d t} &= \grad \mb f(\mb z, t) \cdot \dot {\mb
  z} +
  \frac{\partial \mb f(\mb z, t)}{\partial t} \\
  &= \grad \mb f(\mb z, t) \cdot \mb f(\mb z, t) +
  \frac{\partial \mb f(\mb z, t)}{\partial t} \label{eq:totalderiv}
\end{align}
Well conditioned flows will place constant, or nearly constant, force on
particles as they travel. 
Thus, in this work we propose regularizing the dynamics with two penalty terms,
one term regularizing $\mb f$ and the other $\grad \mb f$.
The first
penalty, presented in Section \ref{sec:ot}, is a measure of the distance
travelled under the flow $\mb f$, and can alternately be interpreted as the kinetic
energy of the flow. This penalty term is based off of numerical methods in optimal
transport, and encourages particles to travel in straight lines with constant
speed. The second penalty term, discussed in Section \ref{sec:jf}, performs
regularization on the Jacobian of the vector field. Taken together the two
terms ensure that the force experienced by a particle under the flow is constant
or nearly so.

These two regularizers will promote dynamics
that follow numerically easy-to-integrate paths, thus greatly speeding up training time.

\section{Optimal transport maps \& Benamou-Brenier}\label{sec:ot}
There is a remarkable similarity between density estimation using continuous
time normalizing flows, and the
calculation of the optimal transport map between two densities using the
Benamou-Brenier formulation 
\citep{benamou2000computational,santambrogio2015optimal}.
While a review of optimal transport theory is far outside the scope of this paper,
here we provide an informal summary of key ideas relevant to continuous
normalizing flows.
The quadratic-cost optimal transport map between two densities $p(\mb x)$ and
$q(\mb x)$ is a map
$\mb z:\Rd\mapsto \Rd$ minimizing the transport cost
\begin{equation}\label{eq:quadot}
  M(\mb z) = \int \| \mb x - \mb z(\mb x) \|^2 p(\mb x) \, \mathrm d \mb x
\end{equation}
subject to the constraint that $\int_A q(\mb z) \,\mathrm d\mb z = \int_{\mb
z^{-1}(A)} p(\mb x)
\, \mathrm d\mb x$, in other words that the measure of any set $A$ is preserved under
the map $\mb z$.
In a seminal work, \citet{benamou2000computational} showed that rather than
solving for minimizers of \eqref{eq:quadot} directly, an indirect (but
computationally efficient) method is available by writing $\mb z(\mb x, T)$ as the solution
map of a flow under a vector field $\mb f$ (as in \eqref{eq:dynamics}) for time
$T$, by minimizing 
\begin{subequations}
\begin{alignat}{2}
&\!\min_{\mb f, \rho}        &\qquad& \int_0^T \int \| \mb f(\mb x, t) \|^2  \rho_t(\mb
  x) \, \mathrm d \mb x \mathrm d t
\label{eq:bb}\\
&\text{subject to} &      & \frac{\partial\rho_t}{\partial t}  =
-\div{\rho_t
  \mb f },\label{eq:continuity}\\
  &                  &      & \rho_0(\mb x) = p,\label{eq:constraint1} \\
  &                  &      & \rho_T(\mb z) = q.\label{eq:constraint2}
\end{alignat}\label{eq:bbopt}
\end{subequations}
The objective function \eqref{eq:bb} is a measure of the \emph{kinetic energy}
of the flow. The constraint \eqref{eq:continuity} ensures probability mass is
conserved. The latter two constraints guarantee the learned distribution agrees
with the source $p$ and target $q$.
  Note that the kinetic energy \eqref{eq:bb} is an upper bound on the transport
  cost, with equality only at optimality.

The optimal flow $\mb f$ minimizing \eqref{eq:bbopt} has several particularly appealing
properties. First, particles induced by the optimal flow $\mb f$ \emph{travel in
straight lines}. Second, particles travel with \emph{constant speed}. Moreover, under suitable conditions on the source and target distributions, the optimal solution map is unique \cite{villani2008optimal}.
Therefore the solution map $\mb z(\mb x,t)$ is entirely characterized by the initial
and final positions:  $\mb z(\mb x,t) = (1-\frac{t}{T}) \mb z(\mb x, 0) +
\frac{t}{T} \mb z(\mb x,
T)$. Consequently, given an optimal $\mb f$ it is extraordinarily easy to solve
\eqref{eq:dynamics} numerically with minimal computational effort.

\subsection{Linking normalizing flows to optimal transport}\label{sec:link}
Now suppose we wish to minimize \eqref{eq:bb}, with $q(\mb z)$ a unit normal distribution,
and $p(\mb x)$ a data distribution,
unknown to us, but from which we have drawn $N$ samples, and which we model as a
discrete distribution of Dirac masses.
Enforcing the initial condition is trivial because we have sampled
from $p$ directly.
The continuity equation \eqref{eq:continuity} need not be enforced because we are tracking a finite
number of sampled particles.
However the final time condition $\rho_T = q$ cannot be implemented
directly, since we do not have direct control on the form $\rho_T(\mb z)$
takes. Instead, introduce a Kullback-Leibler term
to \eqref{eq:bb} penalizing discrepancy between $\rho_T$ and $q$. This penalty
term has an elegant simplification when $p(x)$ is modeled as a distribution of
a finite number of masses, as is done in generative modeling. Setting $\rho_0 =
p_\theta$ a brief derivation yields
\begin{align} 
  \KL(\rho_T || q) &= -\frac{1}{N}\sum_{i=1}^N \log p_{\theta}(\mb x_i) 
\end{align}
 With this simplification
\eqref{eq:bb} becomes
\begin{align}\label{eq:bbreg}
  \begin{split}
    J_\lambda(\mb f) = \frac{\lambda}{N} & \sum_{i=1}^N \int_0^T \| \mb f(\mb z_i, t) \|^2  \mathrm d t \\
    & - \frac{1}{N} \sum_{i=1}^N  \log p_{\theta}(\mb x_i)  
  \end{split}
\end{align}
For further details on this derivation consult the supplementary materials.

The connection between the Benamou-Brenier formulation of the optimal transport
problem on a discrete set of points and continuous normalizing flows is apparent:
the optimal transport problem \eqref{eq:bbreg} is a regularized form of the continuous
normalizing flow optimization problem \eqref{eq:opt1}. We therefore expect that adding a kinetic
energy regularization term to FFJORD will encourage solution trajectories to
prefer straight lines with constant speed.

\begin{algorithm}[tb]
   \caption{RNODE: regularized neural ODE training of FFJORD}
   \label{alg:rnode}
\begin{algorithmic}
  \STATE {\textbf{Input}: data $X = \{\mb x_i\}$, $i=1,\cdots,N$, dynamics
  $\mb f(\cdot\,; \theta)$, \hspace{3em}final time $T$, regularization strength $\lambda_J$
and $\lambda_K$}
\STATE{\textbf{initialize} $\theta$}
  \WHILE{$\theta$ not converged}
  \STATE{Sample $\mb \epsilon$ from standard normal distribution}
  \STATE{Sample minibatch $\{\mb x_j\}$ of size $m$ from $X$}
  \STATE{Set $\mb z_j(0) = \mb x_j$, $l_j(0)=E_j(0)=n_j=0$ }
  \STATE{Numerically solve up to time $T$ the system 
      \begin{equation}
        \begin{cases}
        \dot{\mb z}_j = \mb f(\mb z_j, t;\theta) \\
        \dot{l}_j = \mb \epsilon^\trp \grad \mb f(\mb z_j, t;\theta) \mb\epsilon \\
      \dot{E}_j = \norm{\mb f(\mb z_j, t;\theta) \mb}^2 \\
    \dot{n}_j = \norm{\epsilon^\trp \grad \mb f(\mb z_j, t;\theta)}^2 
        \end{cases} \nonumber
      \end{equation}
      }
      \STATE{Compute  \begin{align*}L(\theta)=\frac{1}{m}\sum_{j=1}^m &-\log q(z_j(T)) -
          l_j(T) \\&+
      \lambda_J n_j(T) + \lambda_K E_j(T) \end{align*}}
      \STATE{Compute $\grad_\theta L(\theta)$ using the adjoint sensitivity
    method by numerically solving the adjoint equations}
      \STATE{Update $\theta \gets \theta - \tau \grad_\theta L(\theta)$}
  \ENDWHILE
\end{algorithmic}
\end{algorithm}
\section{Unbiased Frobenius norm regularization of the Jacobian}\label{sec:jf}

\begin{figure}[t]
\vskip 0.2in
\begin{center}
  \centerline{\includegraphics[width=2.5in]{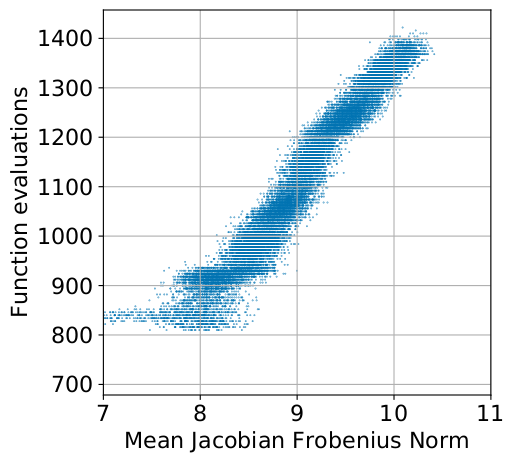}}
\caption{Number of function evaluations vs Jacobian Frobenius norm of flows on
CIFAR10 during training with vanilla FFJORD, using an adaptive ODE solver.}
\label{fig:fe-vs-jf}
\end{center}
\vskip -0.2in
\end{figure}

Refering to equation \eqref{eq:totalderiv}, one can see that even if $\mb f$ is regularized to be small, via a kinetic energy penalty term, if the Jacobian is large then the force experienced by a particle may also still be large. As a result, the error of the numerical integrator can be large, which may lead an adaptive solver to make many function evaluations. 
This relationship is apparent in Figure \ref{fig:fe-vs-jf}, where we empirically
demonstrate the correlation between the number of function evaluations of
$\mb f$ taken
by the adaptive solver, and the size of the Jacobian norm of $\mb f$. The correlation is
remarkably strong: dynamics governed by a poorly conditioned Jacobian matrix require the adaptive solver to take many small time steps.

Moreover, in particle-based methods, the kinetic energy term forces dynamics to travel in straight lines \emph{only on data seen during training}, and so the regularity of the map is only guaranteed on trajectories taken by training data. The issue here is one of generalization: the map may be irregular on off-distribution or perturbed images, and cannot be remedied by the kinetic energy term during training alone. In the context of generalization, Jacobian regularization is analagous to gradient regularization, which has been shown to improve generalization \cite{lecun92,novak18}.

For these reasons, we also propose regularizing the Jacobian through its Frobenius norm.
The Frobenius norm $\|\cdot\|_F$ of a real matrix $A$ can be thought of as the
$\ell_2$ norm of the matrix $A$ vectorized
\begin{equation}
  \|A\|_F = \sqrt{\sum_{i,j} a_{ij}^2 }
\end{equation}
Equivalently it may be computed as
\begin{equation}
  \|A\|_F = \sqrt{\tr(AA^\trp ) }
\end{equation}
and is the Euclidean norm of the singular values of a matrix.
In trace form, the Frobenius norm lends itself to estimation using a Monte-Carlo trace
estimator \citep{hutchinson1990stochastic,avrom11}. For real matrix $B$, an unbiased estimate of the trace is given by
\begin{equation}
  \tr(B) = \E_{\mb \epsilon\sim \mathcal N(0,1)} \left[\mb \epsilon^\trp
  B \mb \epsilon\right]
\end{equation}
where $\mb \epsilon$ is drawn from a unit normal distribution. Thus the squared
Frobenius norm can be easily estimated by setting $B=AA^\trp $.

Turning to the Jacobian $\grad \mb f(\mb z)$ of a vector valued function
$\mb f:\Rd \mapsto \Rd$, recall that the vector-Jacobian product $\mb
\epsilon^\trp \grad \mb f(\mb z)$ may be quickly
computed through reverse-mode automatic differentiation. Therefore an unbiased
Monte-Carlo estimate of the Frobenius norm of the Jacobian is readily available
\begin{align}
  \|\grad \mb f(\mb z) \|_F^2 &= \E_{\epsilon \sim \mathcal N(0,1) }
  \mb \epsilon^\trp \grad \mb f(\mb z) \grad \mb f(\mb z)^\trp \mb \epsilon \\
 &=  \E_{\epsilon \sim \mathcal N(0,1) }
  \|\mb \epsilon^\trp \grad \mb f(\mb z) \| ^2
\end{align}
Conveniently, in the FFJORD framework the quantity $\mb \epsilon^\trp \grad
\mb f(\mb z)$ must be computed during the estimate of the probability
distribution under the flow, in the Monte-Carlo estimate of the divergence term
\eqref{eq:divtrest}. Thus Jacobian Frobenius norm regularization is
available with essentially \emph{no extra computational cost}.
 
\begin{table*}[t]
   \caption{Log-likelihood (in bits/dim) and training time (in hours) on
     validation images with uniform dequantization. Results on clean images are
     found in the supplemental materials. For comparison we report both
 the results of the original FFJORD paper \citep{grathwohl_ffjord} and our own independent run of FFJORD
 (``vanilla") on CIFAR10 and MNIST. Vanilla FFJORD did not train on
 ImageNet64 (denoted by ``x'').  Also reported are results for other flow-based generative
 modeling papers. Our method (FFJORD with RNODE) has comparable log-likelihood
 as FFJORD but is significantly faster.}
   \label{tab:results}
   \vskip 0.15in
   \begin{center}
   \begin{small}
     \begin{sc}
     \begin{adjustbox}{max width=\textwidth}
   \begin{tabular}{lllcllcllcll}
     \toprule
      & \multicolumn{2}{c}{MNIST} & & \multicolumn{2}{c}{CIFAR10} & &
      \multicolumn{2}{c}{ImageNet64} & & \multicolumn{2}{c}{CelebA-HQ256} \\
      & bits/dim &  time & & bits/dim & time & & bits/dim & time \\
  \midrule
     FFJORD, original  & 0.99  &  -   & &
     3.40  & $\geq$ 5 days && - & -&& - & - \\
     FFJORD, vanilla               & 0.97 & 68.5 & & 3.36 & 91.3  && x  &  x
     && - & -\\
     FFJORD RNODE (ours)           & 0.97 & 24.4  & & 3.38 & 31.8  && 3.83
     & 64.1 && 1.04 &  6.6 days\\

     \midrule
     RealNVP \scriptsize{\cite{realnvp}}      & 1.06  &  -    & & 3.49  & - &&
     3.98 & -&& - & -\\
     i-ResNet \scriptsize{\cite{iresnet}}& 1.05 &  -    & & 3.45 & -  && - & -
     && - & -\\
     Glow \scriptsize{\cite{glow}}      & 1.05  &   -   & & 3.35  & - && 3.81 &
     - && 1.03 & 7 days\footnotemark\\
     Flow++ \scriptsize{\cite{flow++}}      & - &   -   & & 3.28   &-  &&
     -  & -&& - & - \\
     Residual Flow \scriptsize{\cite{behrmann2018invertible}}& 0.97 &    -  & &
     3.28 & - && 3.76 & - && 0.99 & -\\
        \bottomrule
     \end{tabular}
     \end{adjustbox}
   \end{sc}
     \end{small}
     \end{center}
     \vskip -0.1in
 \end{table*}
\begin{figure*}
  \vskip 0.2in
  \centering
    \begin{subfigure}[t]{.45\textwidth}
    \centering
    \includegraphics[width=3in]{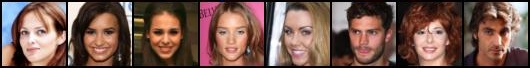}
    \caption{real}\label{fig:realceleb64}
    \end{subfigure} 
    \hspace{3em}
    \begin{subfigure}[t]{.45\textwidth}
    \centering
    \includegraphics[width=3in]{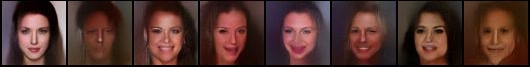}
    \caption{generated}\label{fig:fakeceleb64}
    \end{subfigure}
    \caption{Quality of generated samples samples on 5bit CelebA-HQ64 with RNODE. Here
      temperature annealing \cite{glow} with $T=0.7$ was used to generate
    visually appealing images. For full sized CelebA-HQ256 samples, consult the
  supplementary materials.}\label{fig:celebahq64}
    \vskip -0.2in
\end{figure*}

\section{Algorithm description}
All together, we propose modifying the objective function of the FFJORD continuous normalizing flow
\cite{grathwohl_ffjord} with the two regularization penalties of Sections
\ref{sec:ot} \& \ref{sec:jf}. The proposed method is called RNODE, short for
regularized neural ODE.  Pseudo-code of the method
is presented in Algorithm \ref{alg:rnode}. The optimization problem to be solved is
\begin{align}
  \min_{\mb f} \frac{1}{N d} \sum_{i=1}^N &-\log q(\mb z(\mb x_i, T))
  \nonumber\\ 
  - &\int_0^T
  \div{\mb f}(\mb z(\mb x_i, s), s) \,\mathrm d s \nonumber \\
  + &\lambda_K \int_0^T \norm{\mb f (\mb z(\mb x_i, s), s)}^2 \,\mathrm d s \nonumber \\
  + &\lambda_J \int_0^T \norm{\grad_{\mb z} \mb f (\mb z(\mb x_i, s), s)}^2_F \,\mathrm
  d s  \label{eq:objective}
  \end{align}
where $\mb z(\mb x, t)$ is determined by numerically solving
\eqref{eq:dynamics}. Note that we take the mean over number of samples
\emph{and} input dimension. This is to ensure that the choice of regularization
strength $\lambda_K$ and $\lambda_J$ is independent of dimension size and sample
size.

To compute the three integrals and the log-probability under $q$ of $\mb z(\mb
x,T)$ at final time $T$, we augment the dynamics of the ODE with three
extra terms, so that the entire system solved by the numerical integrator is
\begin{equation}
  \begin{cases}
    \dot{\mb z} = \mb f(\mb z, t) \\
    \dot{l} = \div{\mb f}(\mb z, t) \\
    \dot{E} = \norm{\mb f(\mb z, t)}^2 \\
    \dot{n} = \norm{\grad \mb f(\mb z, t)}^2_F \\
    \mb z(0) = \mb x$,\quad $E(0)=l(0)=n(0)=0
  \end{cases} \tag{RNODE}
\end{equation}
Here $E$, $l$, and
$n$ are respectively the kinetic energy, the log determinant of the Jacobian,
and the integral of the Frobenius norm of the Jacobian.

Both the divergence term and the Jacobian Frobenius norm are approximated with
Monte-Carlo trace estimates. In our implementation, the Jacobian Frobenius
estamate reuses the computatian $\mb \epsilon^\trp \grad f$ from the divergence
estimate for efficiency. We remark that the kinetic energy term only requires
the computation of a dot product. Thus just as in FFJORD, our implementation
scales linearly with the number of time steps taken by the ODE solver.

Gradients of the objective function with respect to
the network parameters are computed using the adjoint sensitivity method
\cite{pontryagin1962mathematical,chen2018neural}.

 \addtocounter{footnote}{0}
 \footnotetext[2]{\cite{glow} trained with 40 GPUs for a
week; in contrast we train with four GPUs in just under a week.}
\section{Experimental design}
\begin{figure*}[ht]
  \vskip 0.2in
    \centering
    \begin{subfigure}[t]{.33\textwidth}
      \includegraphics[height=2in]{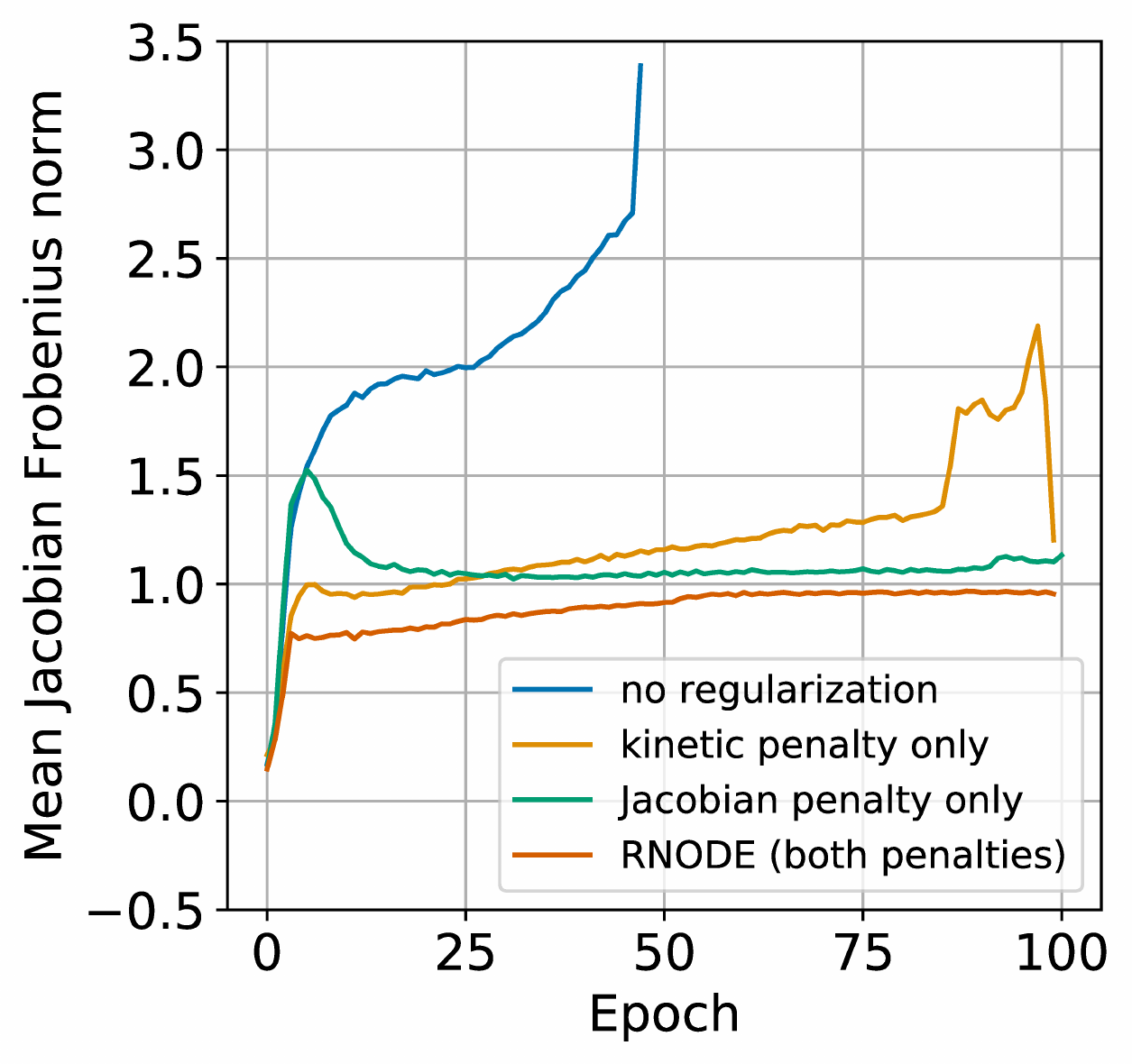}
    \caption{Jacobian norm}\label{fig:jf}
    \end{subfigure}
    \begin{subfigure}[t]{.33\textwidth}
      \includegraphics[height=2in]{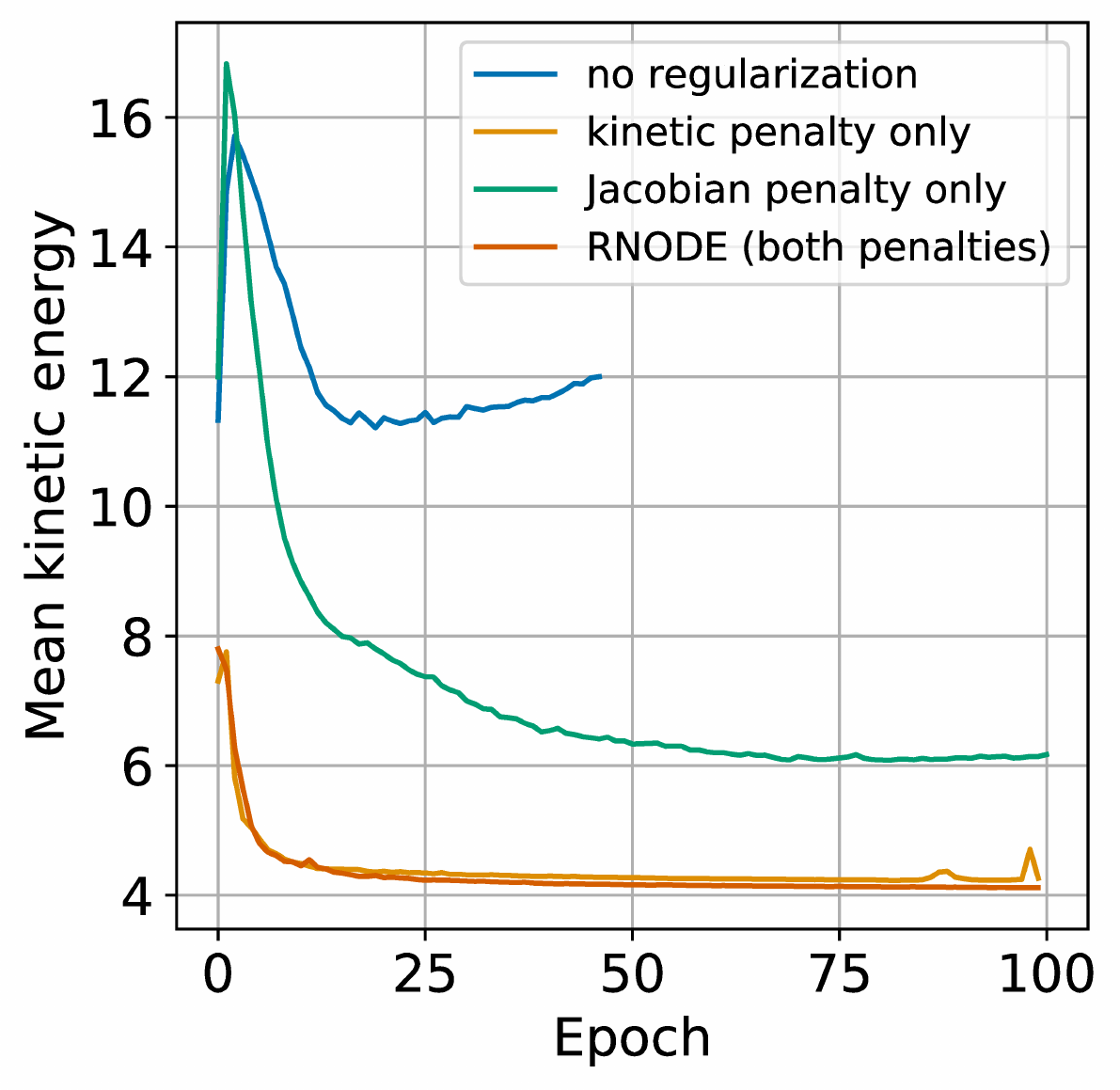}
    \caption{Kinetic energy}\label{fig:ke}
  \end{subfigure}
    \begin{subfigure}[t]{.33\textwidth}
      \includegraphics[height=2in]{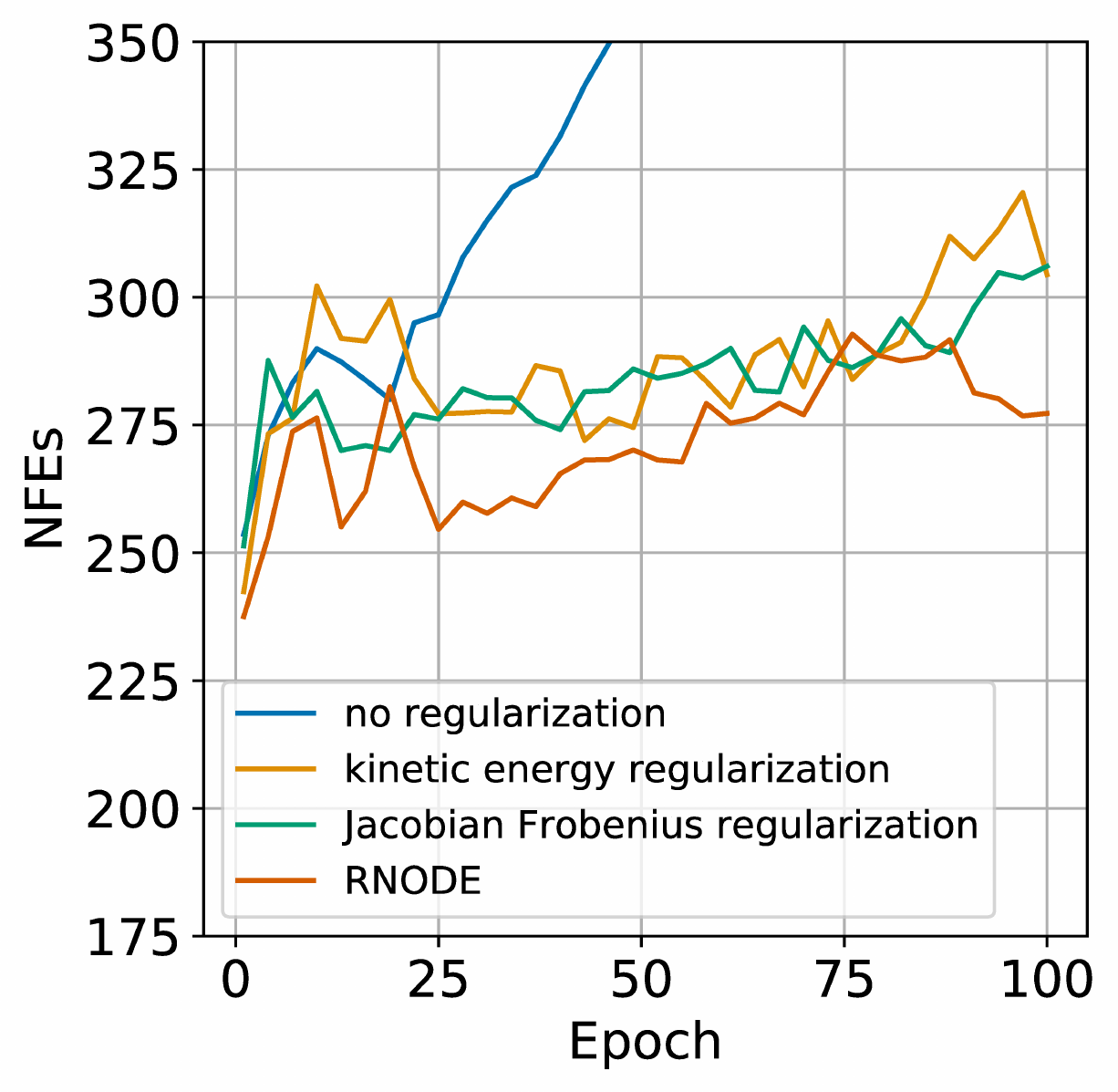}
    \caption{Function evaluations}\label{fig:nfes}
  \end{subfigure}
  \caption{Ablation study of the effect of the two regularizers, comparing two measures of flow regularity during training with a fixed
    step-size ODE solver.  Figure
  \ref{fig:jf}: mean Jacobian Frobenius norm as a function of training
epoch.  Figure \ref{fig:ke}: mean kinetic energy of the flow as a function of
training epoch. Figure \ref{fig:nfes}: number of function evaluations.}\label{fig:ablation}
\end{figure*}
Here we demonstrate the benefits of regularizing neural ODEs on generative
models, an application where neural ODEs have shown strong empirical
performance. We use four datasets: CIFAR10 \citep{cifar},  MNIST \cite{mnist},
downsampled ImageNet (64x64) \cite{imagenet64}, and 5bit CelebA-HQ (256x256) \cite{celebahq}. We use an 
identical neural architecture to that of \citet{grathwohl_ffjord}. 
The dynamics are defined by a neural network $\mb f(\mb z,t;\mb
\theta(t)): \Rd\times \mathbb{R}_+ \mapsto \Rd$ where $\mb \theta(t)$ is piecewise
constant in time.  On MNIST we use
10 pieces; CIFAR10 uses 14; downsampled ImageNet uses 18; and CelebA-HQ uses 26
pieces. Each piece is a 4-layer deep convolutional network
comprised of 3x3
kernels and softplus activation functions. Intermediary layers have 64 hidden
dimensions, and time $t$ is concatenated to the spatial input $z$. The
integration time of each piece is $[0,1]$. Weight matrices
are chosen to imitate the multi-scale architecture of Real NVP \citep{realnvp},
in that images are `squeezed' via a permutation to halve image height
and width but quadruple the number of channels. 
Divergence of $\mb f$ is estimated using the Gaussian Monte-Carlo trace
estimator with one sample of
fixed noise per solver time-step.

On MNIST and CIFAR10 we train with a batch size of 200 and train for 100 epochs on a single GPU\footnote{GeForce RTX 2080 Ti}, using the Adam optimizer \citep{adam} with a
learning rate of \num{1e-3}. On the two larger datasets, we train with four
GPUs, using
a per-GPU batch size of respectively 3 and 50 for CelebA-HQ and ImageNet.  Data is preprocessed by perturbing with uniform
noise followed by the logit transform.

The reference implementation of FFJORD solves the dynamics using a Runge-Kutta 4(5) adaptive solver \cite{dormand1980family} with error tolerances \num{1e-5}
and initial step size \num{1e-2}. We have found that using less accurate solvers
on the reference implementation of FFJORD results in numerically unstable
training dynamics. In contrast, a
simple fixed-grid four stage Runge-Kutta solver suffices for RNODE during
training on MNIST and CIFAR10, using a step size of
$0.25$. The step size was determined based on a simple heuristic of starting
with $0.5$ and decreasing the step size by a factor of two until the discrete
dynamics were stable and achieved good performance. The Runge-Kutta 4(5)
adaptive solver
was used on the two larger datasets. We have also observed that
RNODE improves the training time of the adaptive solvers as well, requiring many
fewer function evaluations; however in Python we have found that the fixed grid
solver is typically quicker at a specified number of function evaluations. At
test time RNODE uses the same adaptive solver as FFJORD.

We always initialize RNODE so that $\mb f(z,t)=0$; thus training
begins with an initial identity map. This is done by zero-ing the parameters of
the last layer in each piece (block), following \citet{goyal17}. The identity map is an appropriate choice
because it has zero transport cost and zero
Frobenius norm. Moreover the identity map is trivially solveable for any
numerical solver, thus training begins without any effort required on the
solver's behalf.

On all datasets we set both the kinetic energy regularization
coefficient $\lambda_K$ and the Jacobian norm coefficient $\lambda_J$ to 0.01.

\section{Results}
\begin{figure*}
  \vskip 0.2in
  \centering
    \begin{subfigure}[t]{.45\textwidth}
    \centering
    \includegraphics[width=2.5in]{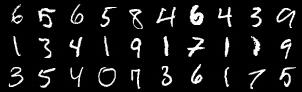}
    \caption{real MNIST images}\label{fig:realmnist}
    \end{subfigure} 
        \begin{subfigure}[t]{.45\textwidth}
    \centering
    \includegraphics[width=2.5in]{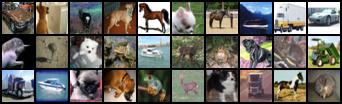}
    \caption{real CIFAR10 images}\label{fig:realcifar10}
    \end{subfigure} 
        \begin{subfigure}[t]{.45\textwidth}
    \centering
    \includegraphics[width=2.5in]{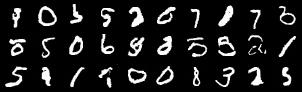}
    \caption{vanilla FFJORD}\label{fig:munreg}
    \end{subfigure}
    \begin{subfigure}[t]{.45\textwidth}
    \centering
    \includegraphics[width=2.5in]{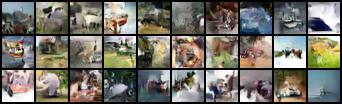}
    \caption{vanilla FFJORD}\label{fig:cunreg}
    \end{subfigure}
        \begin{subfigure}[t]{.45\textwidth}
    \centering
    \includegraphics[width=2.5in]{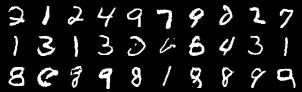}
    \caption{FFJORD RNODE}\label{fig:mreg}
    \end{subfigure}
    \begin{subfigure}[t]{.45\textwidth}
    \centering
    \includegraphics[width=2.5in]{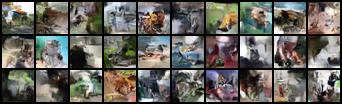}
    \caption{FFJORD RNODE}\label{fig:creg}
    \end{subfigure}
    \caption{Quality of generated samples samples with and without
    regularization on MNIST, left, and CIFAR10, right.}\label{fig:gen}
    \vskip -0.2in
\end{figure*}
A comparison of RNODE against FFJORD and other flow-based generative models is presented in Table \ref{tab:results}. We
report both our running of ``vanilla" FFJORD and the results as originally reported
in \citep{grathwohl_ffjord}. We highlight that RNODE  runs roughly 2.8x faster than FFJORD on both datasets, while achieving or surpassing the performance of FFJORD. This can further be seen in
Figure \ref{fig:bpd} where we plot bits per dimension ( $-\frac{1}{d} \log_2 p(x)$, a normalized measure of
log-likelihood) on the validation set as a function of training epoch, for both
datasets. Visual inspection of the sample quality reveals no qualitative difference between regularized and unregularized approaches; refer to Figure \ref{fig:gen}. 
Generated images for downsampled ImageNet and CelebA-HQ are deferred to the
supplementary materials; we provide smaller generated images for networks
trained on CelebA-HQ 64x64 in
Figure \ref{fig:celebahq64}.

Surprisingly, our run of ``vanilla" FFJORD achieved slightly better performance than the results reported in \cite{grathwohl_ffjord}.
We suspect the discrepancy in performance and run times between our implementation of FFJORD
and that of the original paper is due to batch size:
\citeauthor{grathwohl_ffjord} use a batch size of 900 and train on six GPUs,
whereas on MNIST and CIFAR10 we use a batch size of 200 and train on a single GPU.

We were not able to train vanilla FFJORD on ImageNet64, due to numerical
underflow in the adaptive solver's time step. This issue cannot be remedied by
increasing the solver's error tolerance, for this would bias the log-likelihood
estimates on validation.

\subsection{Ablation study on MNIST}
In Figure \ref{fig:ablation}, we compare the effect of each regularizer by itself on the training
dynamics with the fixed grid ODE solver on the MNIST dataset. Without any
regularization at all, training dynamics are numerically unstable and fail after
just under 50 epochs. This is precisely when the Jacobian norm grows large;
refer to Figure \ref{fig:jf}. Figure \ref{fig:jf} demonstrates that each
regularizer by itself is able to control the Jacobian norm. The
Jacobian regularizer is better suited to this task, although it is interesting
that the kinetic energy regularizer also improves the Jacobian norm.
Unsurprisingly Figure \ref{fig:ke} demonstrates the addition of the kinetic energy regularizer
encourages flows to travel a minimal distance. In addition, we see that the
Jacobian norm alone also has a beneficial effect on the distance particles
travel. Overall, the results support our theoretical reasoning empirically.

%
%

\section{Previous generative flows inspired by optimal transport}
\citet{zhang2019mongeampere} define a neural ODE flow where the dynamics
  are given as the gradient of a scalar potential function. This interpretation
  has deep connections to optimal transport: the optimal transport map
  is the gradient of a convex potential function. 
  \citet{yang19} continue along these lines, and define an optimal transport
  again as a scalar potential gradient. \citet{yang19} enforce that the learned
  map is in fact an optimal transport map by penalizing their objective function with a term measuring
  violations of the continuity equation. 
\citet{ruthotto19} place generative flows within a broader context of mean field
games, and as an example consider a neural ODE gradient potential flow solving the optimal transport problem
in up to 100 dimensions. We also note the recent work of \citet{twomey19}, who proposed regularizing
neural ODEs with an Euler-step discretization of the kinetic energy term to
enforce `straightness', although connections to optimal transport were not
discussed. 

When a flow is the gradient of a scalar potential, the
change of variables formula \eqref{eq:changeint} simplifies so that the
divergence term is replaced by the Laplacian of the scalar potential.
Although mathematically parsimonious and theoretically well-motivated,
we chose not to implement our flow as the gradient of a scalar
potential function due to computational constraints: such an implementation
would require `triple  backprop' (twice
to compute or approximate the Laplacian, and once more for the parameter gradient).
\citet{ruthotto19} circumvented this problem by utilizing special structural
properties of residual networks to efficiently compute the Laplacian.

\section{Discussion}
In practice, RNODE is simple to implement, and only requires augmenting the
dynamics \eqref{eq:dynamics} with two extra scalar equations (one for the
kinetic energy term, and another for the Jacobian penalty). In the setting of
FFJORD, because we may recycle intermediary terms used in the divergence
estimate, the computational cost of evaluating these two extra equations is
minimal. RNODE introduces two extra hyperparameters related to the strength
of the regularizers; we have found these required almost no tuning.

Although the problem of classification was not considered in this work, we
believe RNODE may offer similar improvements both in training time and the
regularity of the classifier learned. In the classification setting we expect
the computional overhead of calculating the two extra terms should be marginal
relative to gains made in training time.

\section{Conclusion}
We have presented RNODE, a regularized method for neural ODEs. This
regularization approach is theoretically well-motivated, and encourages neural
ODEs to learn well-behaved dynamics. As a consequence, numerical integration of
the learned dynamics is straight forward and relatively easy, which means fewer
discretizations are needed to solve the dynamics. In many circumstances,
this allows for the replacement of adaptive solvers with fixed grid solvers,
which can be more efficient during training.  This leads to a substantial speed
up in training time, while still maintaining the same empirical performance, opening the use of neural ODEs to large-scale applications.

\section*{Acknowledgements}
C. F. and A. O. were supported by a grant from the Innovative Ideas Program of the Healthy
Brains and Healthy Lives initiative (HBHL) through McGill University. 

L. N. was supported by AFOSR MURI FA9550-18-1-0502, AFOSR Grant No. FA9550-18-1-0167, and ONR Grant No. N00014-18-1-2527.

A. O. was supported by the Air Force Office of Scientific Research under award number FA9550-18-1-0167

Resources used in preparing this research were provided, in part, by the 
Province of Ontario, the Government of Canada through CIFAR, and 
companies sponsoring the Vector Institute
(\url{www.vectorinstitute.ai/#partners}).

\bibliography{refs}
\bibliographystyle{icml2020}

\clearpage
\appendix
\section{Details of Section \ref{sec:link}: Benamou-Brenier formulation in Lagrangian coordinates}
The Benamou-Brenier formulation of the optimal transportation (OT) problem in Eulerian coordinates is
\begin{subequations}
\begin{alignat}{2}
&\!\min_{\mb f, \rho}        &\qquad& \int_0^T \int \| \mb f(\mb x, t) \|^2  \rho_t(\mb
  x) \, \mathrm d \mb x \mathrm d t
\label{eq:bb}\\
&\text{subject to} &      & \frac{\partial\rho_t}{\partial t}  =
-\div{\rho_t
  \mb f },\label{eq:continuity}\\
  &                  &      & \rho_0(\mb x) = p,\label{eq:constraint1} \\
  &                  &      & \rho_T(\mb z) = q.\label{eq:constraint2}
\end{alignat}\label{eq:bbopt}
\end{subequations}
The connection between continuous normalizing flows (CNF) and OT becomes transparent once we rewrite \eqref{eq:bbopt} in Lagrangian coordinates. Indeed, for regular enough velocity fields $\mb f$ one has that the solution of the continuity equation \eqref{eq:continuity}, \eqref{eq:constraint1} is given by $\rho_t=\mb z(\cdot,t)\sharp p$ where $\mb z$ is the flow
\begin{equation*}
    \dot{\mb z}(\mb x,t)=\mb f (\mb z(\mb x,t),t),\quad \mb z(\mb x,0)=\mb x.
\end{equation*}
The relation $\rho_t=\mb z(\cdot,t)\sharp p$ means that for arbitrary test function $\phi$ we have that
\begin{equation*}
    \int \phi(\mb x) \rho_t(\mb x,t)d\mb x = \int \phi(\mb z(\mb x,t)) p(\mb x) d\mb x
\end{equation*}
Therefore \eqref{eq:bbopt} can be rewritten as
\begin{subequations}
\begin{alignat}{2}
&\!\min_{\mb f}        &\qquad& \int_0^T \int \| \mb f(\mb z(\mb x,t), t) \|^2  p(\mb
  x) \, \mathrm d \mb x \mathrm d t
\label{eq:bb_Lag}\\
&\text{subject to} &      & \dot{\mb z}(\mb x,t)=\mb f (\mb z(\mb x,t),t),\label{eq:continuity_Lag}\\
  &                  &      &  \mb z(\mb x,0)=\mb x,\label{eq:constraint1_Lag} \\
  &                  &      & \mb z(\cdot,T)\sharp p = q .\label{eq:constraint2_Lag}
\end{alignat}\label{eq:bbopt_Lag}
\end{subequations}

Note that $\rho_t$ is eliminated in this formulation. The terminal condition \eqref{eq:constraint2} is trivial to implement in Eulerian coordinates (grid-based methods) but not so simple in Lagrangian ones \eqref{eq:constraint2_Lag} (grid-free methods). To enforce \eqref{eq:constraint2_Lag} we introduce a penalty term in the objective function that measures the deviation of $\mb z(\cdot,T)\sharp p$ from $q$. Thus, the penalized objective function is
\begin{equation}\label{eq:loss_lambda}
    \int_0^T \int \| \mb f(\mb z(\mb x,t), t) \|^2  p(\mb
  x) \, \mathrm d \mb x \mathrm d t+\frac{1}{\lambda} \KL (\mb z(\cdot,T)\sharp p~||~q),
\end{equation}
where $\lambda>0$ is the penalization strength. Next, we observe that this objective function can be written as an expectation with respect to $\mb x \sim p$. Indeed, the Kullback-Leibler divergence is invariant under coordinate transformations, and therefore
\begin{equation*}
    \begin{split}
        \KL (\mb z(\cdot,T)\sharp p~||~q) =& \KL (p~||~\mb z^{-1}(\cdot,T)\sharp
        q)=\KL (p~||~p_\theta) \\
        =& \E_{\mb x \sim p} \log \frac{p(\mb x)}{p_\theta(\mb x)}\\
        =&\E_{\mb x \sim p} \log p(\mb x)-\E_{\mb x \sim p} \log p_\theta(\mb x)
    \end{split}
\end{equation*}
Hence, multiplying the objective function in \eqref{eq:loss_lambda} by $\lambda$ and ignoring the $\mb f$-independent term $\E_{\mb x \sim p} \log p(\mb x)$ we obtain an equivalent objective function
\begin{equation}\label{eq:loss_lambda_p}
    \E_{\mb x \sim p}\left\{\lambda \int_0^T \| \mb f(\mb z(\mb x,t), t) \|^2\, \mathrm d t-\log p_\theta(\mb x) \right\}
\end{equation}
Finally, if we assume that $\{\mb x_i\}_{i=1}^N$ are iid sampled from $p$, we obtain the empirical objective function
\begin{equation}\label{eq:loss_lambda_emp}
    \frac{\lambda}{N} \sum_{i=1}^N \int_0^T \| \mb f(\mb z(\mb x_i,t), t) \|^2\, \mathrm d t-\frac{1}{N}\sum_{i=1}^N \log p_\theta(\mb x_i)
\end{equation}

\section{Additional results}

Here we present additional generated samples on the two larger datasets
considered, CelebA-HQ and ImageNet64. In addition bits/dim on clean images
are reported in Table \ref{tab:additionalresults}.
\begin{figure*}
  \vskip 0.2in
  \centering
    \begin{subfigure}[t]{.45\textwidth}
    \centering
    \includegraphics[width=2.5in]{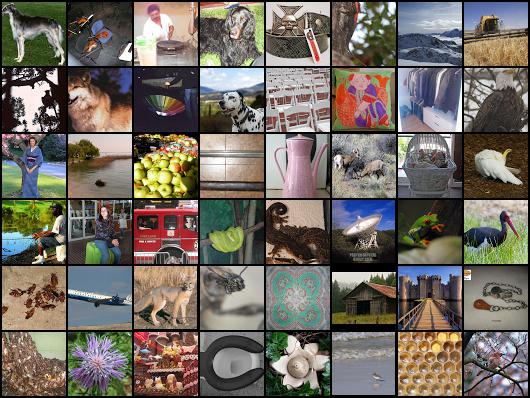}
    \caption{real}\label{fig:realimnet}
    \end{subfigure} 
        \begin{subfigure}[t]{.45\textwidth}
    \centering
    \includegraphics[width=2.5in]{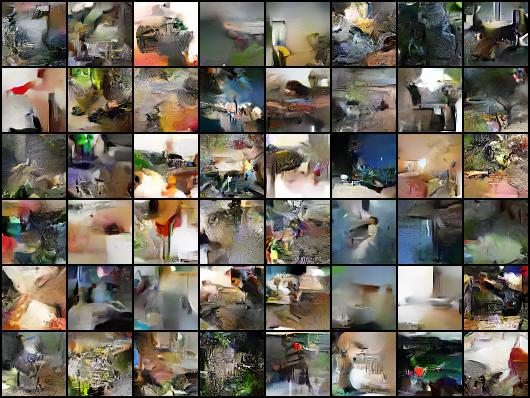}
    \caption{generated}\label{fig:fakeimnet}
    \end{subfigure} 
    \caption{Quality of FFJORD RNODE  generated images on ImageNet-64.}\label{fig:imnetgen}
    \vskip -0.2in
\end{figure*}

\begin{table*}[b]
  \caption{Additional results and model statistics of FFJORD RNODE. Here we
  report validation bits/dim on both validation images, and on validation images 
with uniform variational dequantization (ie perturbed by uniform noise).  We
also report number of trainable model parameters.}
   \label{tab:additionalresults}
   \vskip 0.15in
   \begin{center}
   \begin{small}
   \begin{sc}
   \begin{tabular}{llll}
     \toprule
     dataset & bits/dim (clean) & bits/dim (dirty) & \# parameters \\
  \midrule
  MNIST & 0.92 & 0.97 & \num{8.00e5} \\
  CIFAR10 & 3.25 & 3.38 & \num{1.36e6} \\
  ImageNet64 & 3.72 & 3.83 & \num{2.00e6} \\
  CelebA-HQ256 & 0.72 & 1.04 & \num{4.61e6}\\
        \bottomrule
     \end{tabular}
     \end{sc}
     \end{small}
     \end{center}
     \vskip -0.1in
 \end{table*}
\begin{figure*}
  \vskip 0.2in
  \centering
    \begin{subfigure}[t]{.125\textwidth}
    \centering
    \includegraphics[width=1in]{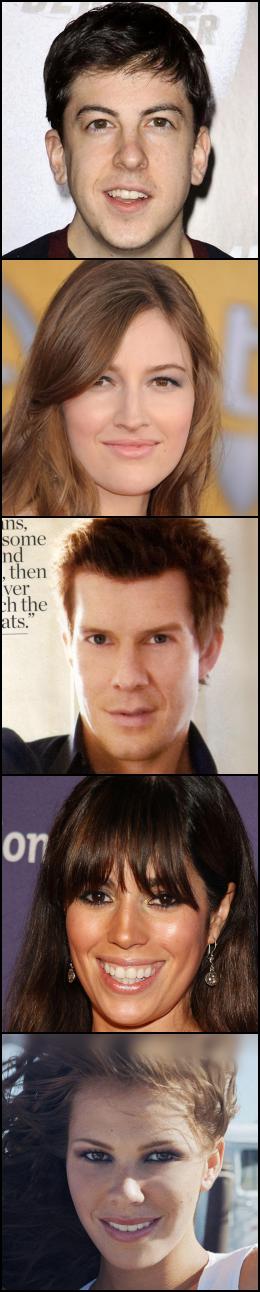}
    \caption{real}\label{fig:realchq}
    \end{subfigure} 
    \hspace{2em}
    \begin{subfigure}[t]{.125\textwidth}
    \centering
    \includegraphics[width=1in]{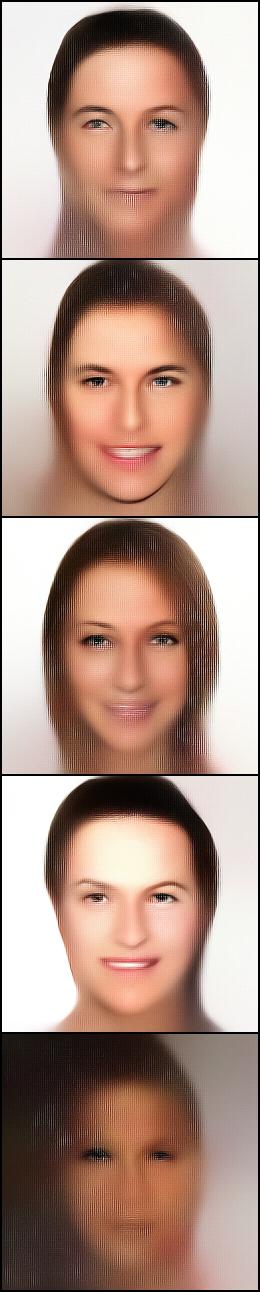}
    \caption{$T=0.5$}\label{fig:T.5chq}
    \end{subfigure} 
    \begin{subfigure}[t]{.125\textwidth}
    \centering
    \includegraphics[width=1in]{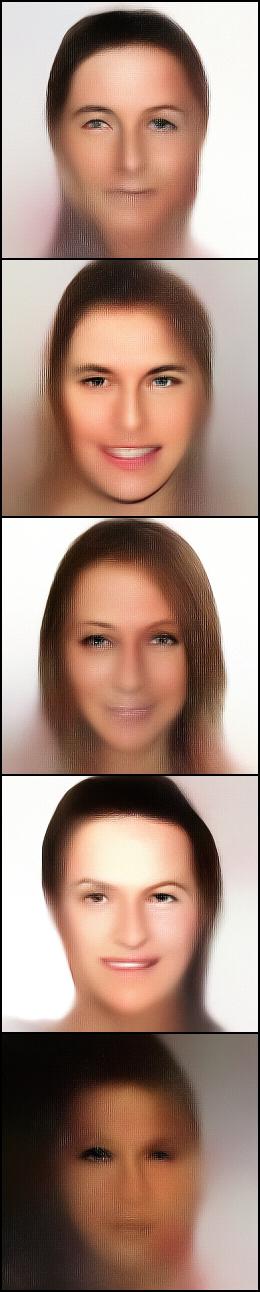}
    \caption{$T=0.6$}\label{fig:T.6chq}
    \end{subfigure} 
    \begin{subfigure}[t]{.125\textwidth}
    \centering
    \includegraphics[width=1in]{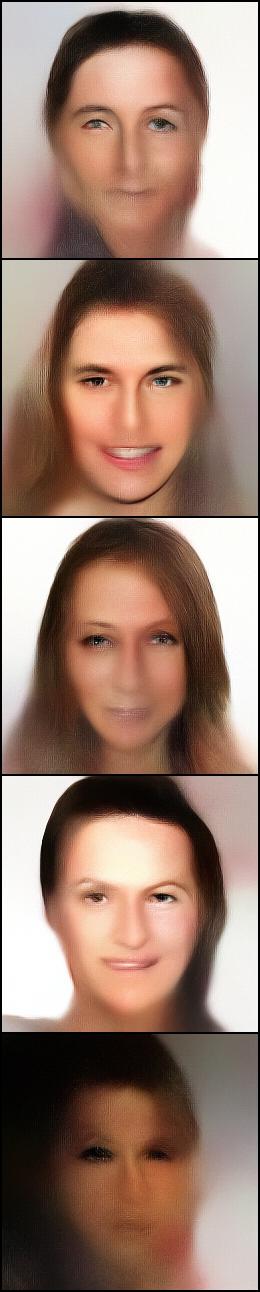}
    \caption{$T=0.7$}\label{fig:T.7chq}
    \end{subfigure} 
    \begin{subfigure}[t]{.125\textwidth}
    \centering
    \includegraphics[width=1in]{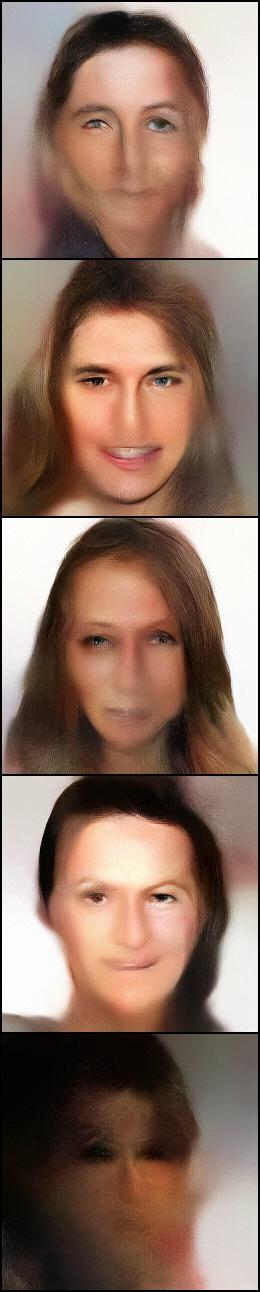}
    \caption{$T=0.8$}\label{fig:T.8chq}
    \end{subfigure} 
    \begin{subfigure}[t]{.125\textwidth}
    \centering
    \includegraphics[width=1in]{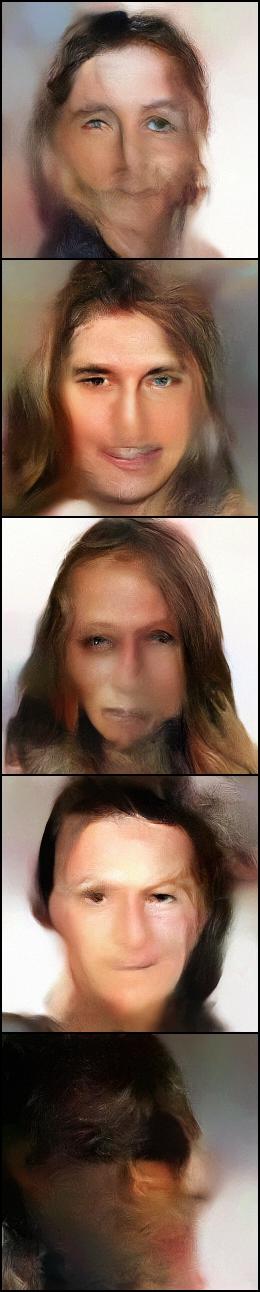}
    \caption{$T=0.9$}\label{fig:T.9chq}
    \end{subfigure} 
    \begin{subfigure}[t]{.125\textwidth}
    \centering
    \includegraphics[width=1in]{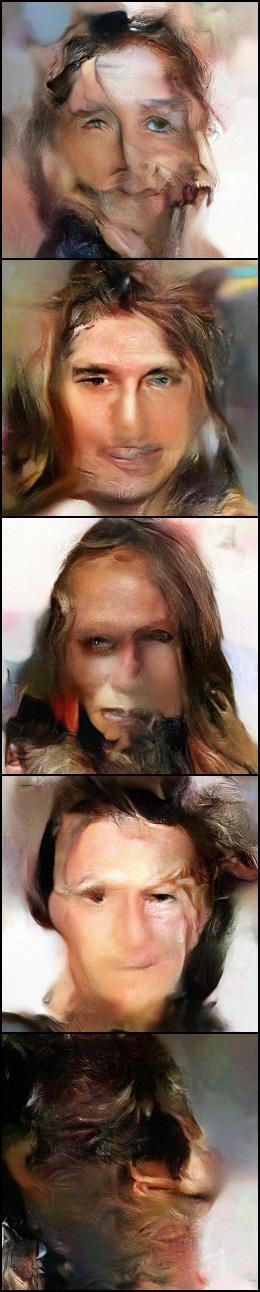}
    \caption{$T=1$}\label{fig:T.10chq}
    \end{subfigure} 
    \caption{Quality of FFJORD RNODE generated images on CelebA-HQ. We use
      temperature annealing,  as described in \cite{glow}, to generate visually appealing
  images, with $T=0.5,\dots,1$.}\label{fig:imnetgen}
    \vskip -0.2in
\end{figure*}

\end{document}